\pdfoutput=1
\documentclass[11pt]{article}
\usepackage{url}
\usepackage{emnlp2021}

\usepackage{times}
\usepackage{latexsym}
\usepackage{graphicx}
\usepackage{amssymb}
\usepackage[T1]{fontenc}
\usepackage[utf8]{inputenc}

\usepackage{microtype}

\title{ParsHate: A Benchmark Dataset for Hate and Target Detection in Persian}

\author{
  Zahra Bokaei \\
  School of Informatics \\
  University of Edinburgh \\
  \texttt{zahra.bokaei@ed.ac.uk}
  \And
  Walid Magdy \\
  School of Informatics \\
  University of Edinburgh \\
  \texttt{wmagdy@ed.ac.uk}
  \And
  Bonnie Webber \\
  School of Informatics \\
  University of Edinburgh \\
  \texttt{bonnie.webber@ed.ac.uk}
}

\usepackage{enumitem}
\usepackage{xcolor}
\newcommand{\change}[2]{#1}
\begin{document}
\maketitle
\begin{abstract}
We introduce ParsHate, a manually annotated dataset of 10,000 Persian tweets spanning 2013–2022, representing the first decade-long benchmark for hate speech detection in Persian. The dataset contains 31\% hateful content and supports both hate detection and multi-label fine-grained target identification across seven structured target categories. ParsHate also distinguishes explicit and implicit hate, marks explicit and implicit targets, and provides span-level rationales. Data collection combines random and score-stratified temporal sampling to reduce keyword-driven bias while preserving natural label distributions. 
Applying SOTA models for Persian hate-speech detection on ParsHate shows moderate performance (79\% F1), especially with samples from earlier years, and low performance with target identification (25.5\% macro-F1). 
This emphasizes the diverse sampling of hate speech in ParsHate and its challenging nature that requires more advanced methods for better performance. Dataset is made publicly available.
\end{abstract}

\section{Introduction}
Hate speech detection has become a central task in Natural Language Processing due to its social impact and the growth of user-generated content \citep{shen2025hatebench}. Over the past decade, many annotated datasets have supported automatic hate speech detection, especially in English and other high-resource languages. These datasets typically label hateful and non-hateful content, and may include hate targets or span-level rationales identifying the evidence behind the label \citep{davidson2017automated, mathew2021hatexplain, kennedy2022introducing}. Some also distinguish explicit from implicit hate, reflecting the complexity of modelling harmful language online \citep{elsherief-etal-2021-latent}. Recent datasets further provide structured target taxonomies and span-level rationales, enabling analysis of both hate expressions and the identity groups involved \citep{mathew2021hatexplain, kennedy2022introducing,elsherief-etal-2021-latent, davidson2017automated}. These schemes provide richer supervision and support more detailed model evaluation.

Despite these advances, Persian remains comparatively under-resourced. To our knowledge, only a limited number of datasets have been developed for Persian hate speech detection, including Pars-Off \citep{ataei2022pars}, PHATE \citep{Delbari_Moosavi_Pilehvar_2024}, PHICAD \citep{davardoustdark}, and Pars-HAO \citep{sheykhlan2023pars}. Although these resources have significantly advanced research in Persian harmful language modeling, they exhibit several structural limitations. Pars-Off relies heavily on keyword-driven sampling and tweets extracted from politically sensitive accounts, which can lead to lexical shortcut learning and under-representation of implicit forms of harassment. PHATE introduces span-level rationales and transformer-based baselines, relies partially on keyword-based candidate selection and covers a  narrow temporal window. In addition, its target annotations are provided in free text and have not been subject to a normalized schema. PHICAD substantially increases the scale of Persian harmful content data, but does not provide a  target taxonomy or distinctions between explicit and implicit hate.

These limitations make it difficult to systematically examine how hate expression evolves across time, how hostility is conveyed implicitly, and how hate targets are represented in Persian online discourse. To address these limitations, we introduce ParsHate,\footnote{\url{https://github.com/zbokaee/ParsHate}} a decade-spanning benchmark dataset of 10,000 manually annotated Persian tweets covering the period from 2013 to 2022, where 31\% of tweets annotated to contain hate-speech. The dataset is constructed using a hybrid temporal sampling strategy that combines fully random sampling with score-stratified sampling based on predictions from a Persian toxicity classifier \citep{bokaei-etal-2025-culture}. This design mitigates keyword-driven sampling bias while ensuring coverage across varying levels of hate without artificially enforcing label balance.

ParsHate supports hate detection and multi-class target identification across seven structured target categories—Gender, Politics, Religion, Nationality, Occupation, Ethnicity, and Other—with fine-grained subcategories. Following prior work on target-aware annotations for implicit hatred \cite{bokaei-etal-2025-culture}, the dataset distinguishes explicit from implicit hate and categorizes implicit expression strategies. It also annotates whether the target itself is expressed explicitly or implicitly and provides span-level rationales highlighting the textual evidence supporting each annotation. These annotations enable further investigation of implicit hate and support the development of more robust multi-label target prediction models. Using ParsHate, we establish benchmark experiments under single year training, integrated multi-year training, and cross-temporal transfer settings to examine how detection performance varies across time and how distributional shifts affect model stability.  Our contributions are as follows:
\begin{itemize}[noitemsep]
    \item Longitudinal Temporal Benchmark: We introduce ParsHate, the first decade-spanning (2013--2022) Persian hate speech dataset and establish cross-year evaluation benchmarks to measure temporal distribution shifts.
    
    \item Structured Target Modeling: We provide fine-grained multi-class target annotations with explicit identity dimensions and explicit/implicit target marking.
    
    \item Implicit--Explicit Hate Annotation: We annotate hate as explicit or implicit, label the strategy used, and provide span-level rationales to support interpretability research.
\end{itemize}

\begin{table*}
    \centering
    \small
    \begin{tabular}{lccccccc}
        Dataset & Source & Period & Size & \% hate & implicit hate & target & implicit target \\
        \hline
        Pars-OFF \cite{ataei2022pars} &  Tweets& 2015-2020 & 10000 &  30\% (OFF)& X  & X & X\\
        PHATE \cite{Delbari_Moosavi_Pilehvar_2024} &  Tweets& 2020-2023 & 7000 &  23\%& X  & \checkmark & X\\
        Pars-HAO \cite{sheykhlan2023pars} & Tweets & unknown & 8000 & 9\% & X & X & X \\
        PHICAD \cite{davardoustdark} & Instagram & unknown & 30000 & 41\% (HOF) & X & X & X \\
        ParsHate (ours) & Tweets & 2013-2022 & 10,000  & 31\% & \checkmark & \checkmark & \checkmark \\
        \hline
    \end{tabular}
    \caption{How ParsHate compares to existing Persian hate-speech datasets}
    \label{tab:PersianDatasets}
\end{table*}

\section{Related Work}
Research on hate speech detection has produced numerous annotated datasets across languages and platforms.  We organise this section around the four dimensions that motivate ParsHate: target taxonomies, implicit hate, span-level rationales, and temporal coverage. This makes the gaps in existing Persian resources explicit and situates ParsHate within the broader dataset landscape.

\textbf{Target Taxonomies:} In English, \citet{mathew2021hatexplain} proposed HateXplain, annotated with class labels and target categories, and \citet{toraman-etal-2022-large} constructed English and Turkish datasets across five target categories. \citet{kennedy2022introducing} introduced the Gab Hate Corpus, a 27K-post dataset with identity-based target categories. In Persian, Pars-OFF \cite{ataei2022pars} provides coarse target types, but not fine-grained identity categories. PHATE \cite{Delbari_Moosavi_Pilehvar_2024} provides free-text target annotations rather than a normalised taxonomy. Pars-HAO \cite{sheykhlan2023pars} and PHICAD \cite{davardoustdark} do not provide structured target schemas.

\textbf{Implicit and Coded Hate:} A growing body of work in English addresses implicit and coded hate, where hostility appears without overt slurs or profanity. \citet{elsherief-etal-2021-latent} introduced a taxonomy of implicit hate with fine-grained categories and target annotations, highlighting the difficulty of modelling indirect hostility. \citet{hartvigsen-etal-2022-toxigen} proposed TOXIGEN, a large-scale machine-generated dataset balanced across identity groups to improve robustness to subtle toxicity. \citet{kennedy2022introducing} additionally annotate an implicitness label in the Gab Hate Corpus, although they do not distinguish whether the target itself is explicitly named or indirectly referenced. Prior work highlights the difficulty of modelling implicit hatred in Persian \citep{bokaei-etal-2025-culture, Delbari_Moosavi_Pilehvar_2024}, yet no existing Persian dataset annotates implicit hate strategies or distinguishes implicit targets. 

\textbf{Span-Level Rationales for Interpretability:}
Span-level rationales support interpretability in English benchmarks such as HateXplain \cite{mathew2021hatexplain}, the Gab Hate Corpus \cite{kennedy2022introducing}, and Latent Hatred \cite{elsherief-etal-2021-latent}. In Persian, only PHATE \cite{Delbari_Moosavi_Pilehvar_2024} provides them.

\textbf{Temporal Coverage and Sampling:}
Studies in high-resource languages show that hateful discourse varies over time \cite{mathew2020hate, garland2022impact}, motivating datasets that support longitudinal analysis. Existing Persian resources, however, cover limited temporal spans: PHATE \cite{Delbari_Moosavi_Pilehvar_2024} covers 2020–2023, Pars-OFF \cite{ataei2022pars} covers 2015–2020, and the temporal windows of Pars-HAO \cite{sheykhlan2023pars} and PHICAD \cite{davardoustdark} are not reported. These datasets also rely on keyword- or profile-based sampling, which introduces lexical bias and under-represents implicit forms of hate.

\textbf{Positioning of ParsHate:} Other widely used English benchmarks, including \cite{tonneau-etal-2024-languages, tonneau2025hateday, rottger-etal-2021-hatecheck, salminen2018anatomy, calabrese2025compositional, ahn2024sharedcon, shen2025hatebench, ghorbanpour2025data}, provide large-scale annotations, target-aware labels, or diagnostic evaluation suites for hate detection. Across the four dimensions above, no existing Persian resource combines structured target taxonomies, implicit-hate strategy annotations, span-level rationales, and longitudinal coverage. ParsHate addresses these gaps by providing a  seven-category target taxonomy with fine-grained subcategories, annotating implicit hate with three rhetorical strategies and implicit target marking, pairing span-level rationales with strategy labels, and covering a decade of Persian hate speech (2013--2022). It also uses score-stratified temporal sampling to reduce keyword-driven bias while preserving natural label distributions. Table~\ref{tab:PersianDatasets} summarises these contributions.

\section{ParsHate: Dataset Construction}
In this section, we describe the construction of ParsHate, a Persian hate speech dataset collected from Twitter. We detail our data retrieval strategy, annotation design, target taxonomy development, and quality control procedures, highlighting the methodological choices made to ensure linguistic diversity, target coverage, and reliable labeling.

\subsection{Data Collection}
To construct ParsHate, we extracted Persian tweets from a Twitter archive containing a 1\% daily sample of the Twitter stream between 2013 and 2022. The archive included approximately 47 million Persian tweets. After removing retweets, duplicates, tweets shorter than five tokens, and tweets containing only URLs, mentions, or hashtags, the corpus was reduced to 29 million tweets. This filtered pool was used for the hybrid temporal sampling procedure described below.

%To reduce noise, we applied several preprocessing steps: removal of retweets and duplicate tweets; exclusion of tweets shorter than five tokens to eliminate low-information instances; and exclusion of tweets containing only URLs, mentions, or hashtags without textual content.

To enable systematic sampling across different levels of hate likelihood, we applied a Persian hate classifier based on fine-tuned Llama 3 \citep{bokaei-etal-2025-culture}, which showed the SOTA performance on existing Persian hate-speech datasets. The model was trained on hate speech datasets in Persian, Arabic and Indonesian, showing a strong performance in prior evaluations. Importantly, none of the datasets used to train this classifier overlap with the tweets included in ParsHate. The model was applied to the filtered corpus to assign probability scores indicating the likelihood of hateful content. These scores were used solely for stratified sampling and not for automatic labeling.

For each year between 2013 and 2022, we sampled 1,000 tweets using a hybrid strategy to balance representativeness and hate intensity coverage. Specifically, 500 tweets were selected randomly from each year to capture naturally occurring discourse. The remaining 500 tweets were selected using classifier-based stratification, with 50 tweets randomly sampled from each probability interval (0.0–0.1, …, 0.9–1.0), where the probability value indicates the likelihood of a tweet to contain hate-speech. This design ensures inclusion of neutral content, borderline cases, and high-confidence hate instances while mitigating lexical bias and preserving temporal diversity.

% \begin{figure*}[t]
% \centering
% \begin{minipage}{0.48\textwidth}
%     \centering
%     \includegraphics[width=\linewidth]{ann2.pdf}
%     \caption{ Dataset instances with corresponding labels.}
%     \label{fig:cloud}
% \end{minipage}
% \hfill
% \begin{minipage}{0.48\textwidth}
%     \centering
%     \includegraphics[width=\linewidth]{misclassification1.pdf}
%     \caption{Misclassification samples for the best models.}
%     \label{fig:RQ1}
% \end{minipage}
% \end{figure*}

\begin{table}[t]
\centering
\includegraphics[width=\columnwidth]{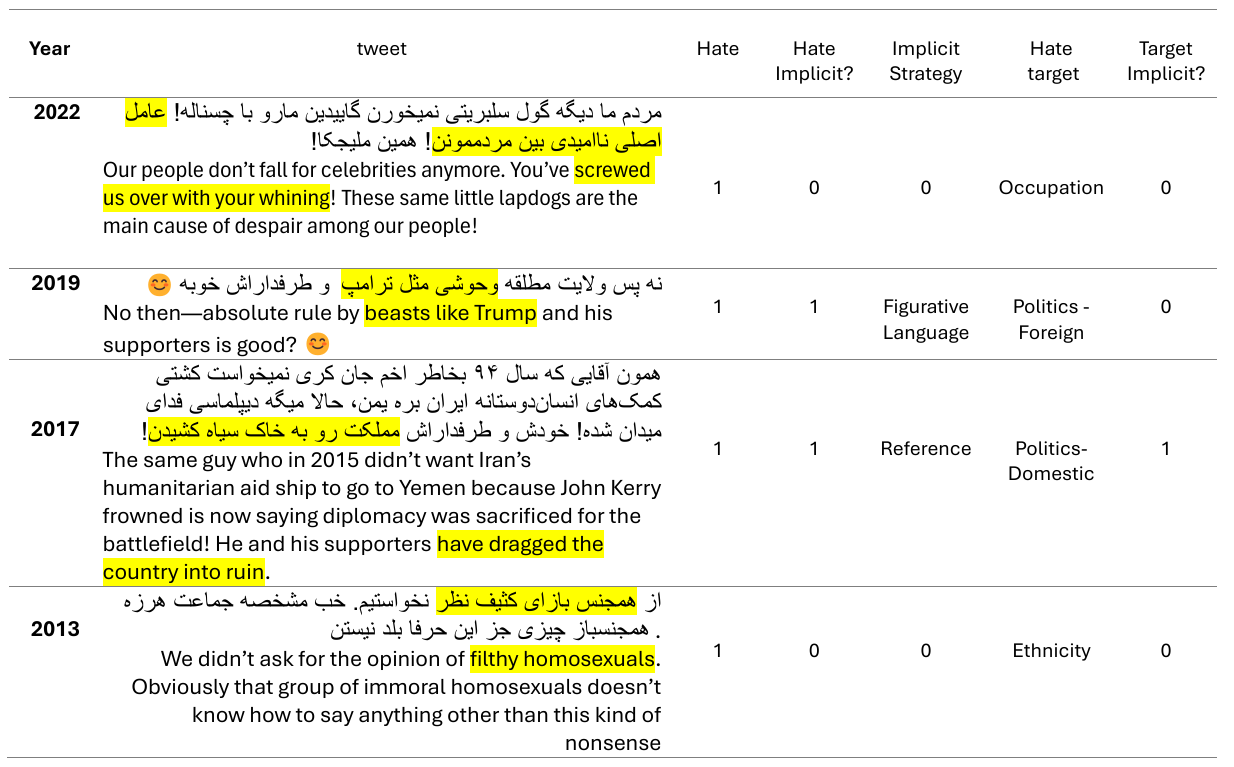}
\caption{Dataset instances with corresponding labels.}
\label{fig:cloud}
\end{table}

\subsection{Annotation Scheme and Guideline}
ParsHate employs a structured annotation framework designed to capture hate presence, rhetorical realization, and identity targets.

\textbf{Hate Definition:} Following \cite{waseem-hovy-2016-hateful}, we define hate as content that expresses or encourages hostility toward groups or individuals based on their association with identity-related categories. This includes negative stereotyping, discrimination, dehumanization, humiliation, or encouragement of harm directed at social or political groups. Each tweet is assigned a binary hate-speech label: hate or non-hate.

\textbf{Target Taxonomy:} The target taxonomy is inspired by identity categories used in the Gab Hate Corpus \citep{kennedy2022introducing}, including gender, political identification, religion, nationality, ethnicity, and other. We extend this framework by adding occupation to reflect identity-based targeting patterns observed in Persian discourse in previous work \cite{Delbari_Moosavi_Pilehvar_2024}. Target annotation allows multi-label assignment when multiple groups are attacked within a single tweet.

\change{Within high-level categories, annotators specified fine-grained subtypes when applicable, following a closed set finalised through the pilot annotation phase and inter-annotator discussion (Appendix~\ref{sec:annotation_guidelines}). Gender distinguishes between \emph{men} and \emph{women}. Religion distinguishes \emph{Islam} from \emph{other religious identities}, reflecting the empirical predominance of Islam-targeted hate in Persian discourse \cite{Delbari_Moosavi_Pilehvar_2024}. Politics distinguishes \emph{domestic} from \emph{foreign} political actors. Nationality distinguishes \emph{Iranian}, \emph{Afghan}, \emph{Arab}, \emph{Western}, and \emph{other}. Occupation, Ethnicity, and the catch-all Other category are not further subdivided. When the target of hate is implied without being explicitly mentioned, it is annotated as an \emph{implicit target}.}{}

\textbf{Implicit Hate:}
Following prior work that distinguishes between explicit and implicit hate \citep{kennedy2022introducing}, we annotate whether the hateful content is expressed explicitly or implicitly.
Implicit hate refers to hostility conveyed without direct slurs or overt statements of hatred, often relying on cultural knowledge, metaphor, sarcasm, insinuation, or exclusionary logic. Based on a pilot annotation phase (250 tweets), we identified three primary strategies of implicit hate:
(i) cultural or historical reference,
(ii) sarcasm or figurative language,
(iii) other indirect rhetorical techniques. Annotators were allowed to assign more than one strategy when multiple rhetorical mechanisms co-occurred within the same tweet. We additionally annotate whether the target itself is explicit or implicit, distinguishing between cases where the attacked group is directly named versus indirectly invoked \cite{kennedy2022introducing}

\begin{figure*}[t]
\centering
\begin{minipage}{0.32\textwidth}
\centering
\includegraphics[width=\linewidth]{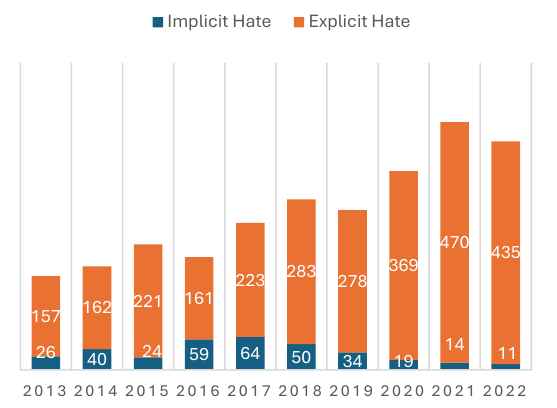}
\caption{Hate Samples Distribution Across years.}
\label{fig:r1}
\end{minipage}
\hfill
\begin{minipage}{0.30\textwidth}
\centering
\includegraphics[width=\linewidth]{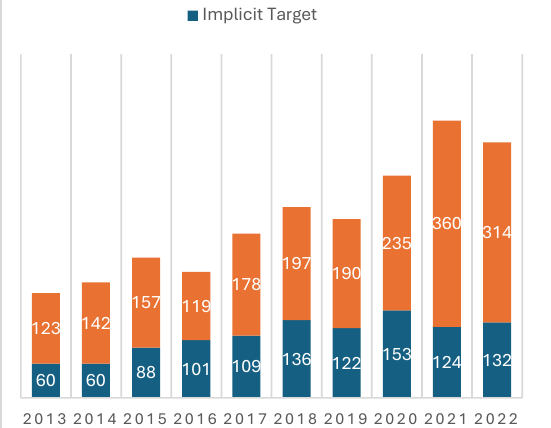}
\caption{Implicit-Explicit Hate Target Distribution Across years.}
\label{fig:r2}
\end{minipage}
\hfill
\begin{minipage}{0.34\textwidth}
\centering
\includegraphics[width=\linewidth]{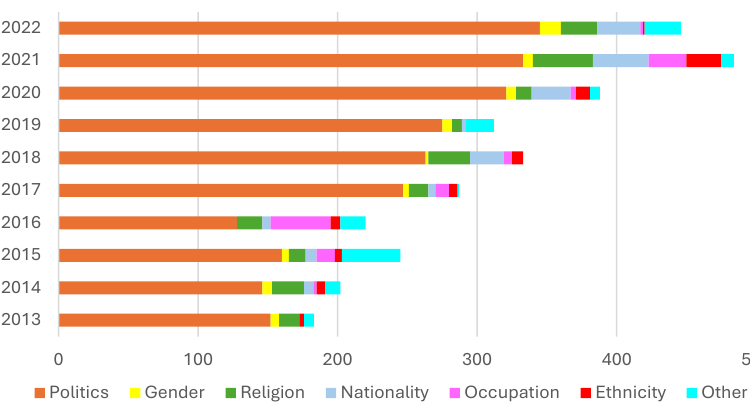}
\caption{Hate Target Distribution Across Years.}
\label{fig:r3}
\end{minipage}
\end{figure*}

\textbf{Span-Level Annotation:} \change{For hate tweets, annotators highlighted the shortest text segment(s) justifying the label. For explicit hate, this was typically overt hate lexicon; for implicit hate, it was the shortest rhetorical device anchoring hostility, such as metaphor, sarcasm, or cultural/historical reference, or the minimal hostile clause when no device was lexically isolable. This follows rationale conventions in HateXplain \citep{mathew2021hatexplain}, the Gab Hate Corpus \citep{kennedy2022introducing}, and Latent Hatred \citep{elsherief-etal-2021-latent}. A 250-tweet pilot aligned annotators on implicit-span cases.}{For hate tweets, annotators highlighted the shortest text segment(s) justifying the assigned labels.}
Span selections were converted into character-level binary labels. A character was retained as part of the final span if at least two annotators marked it; otherwise, it was discarded. Final spans were defined as maximal contiguous sequences of majority-selected characters.

\textbf{Annotation Procedure and Agreement:} All 10{,}000 tweets were independently annotated by three native Persian speakers following detailed written guidelines (Annotator demographics and annotation guidelines are reported in the Appendix.). A pilot phase was conducted to refine definitions, particularly for implicit hate and implicit targeting. \change{Following \citet{rottger-etal-2022-two}, ParsHate uses a \emph{prescriptive} annotation paradigm, with agreement treated as a quality target and disagreements resolved into single instance-level labels through majority voting and discussion-based adjudication.}{} Inter-annotator agreement was measured using Fleiss’ $\kappa$. Agreement scores were $\kappa$ = 0.69 for hate identification, $\kappa$ = 0.60 for target category, $\kappa$ = 0.51 for implicit hate, and $\kappa$ = 0.58 for implicit target. \change{For span annotations, agreement was computed at the character level, yielding $\kappa$ = 0.55 overall, with $\kappa$ = 0.64 for explicit spans and $\kappa$ = 0.44 for implicit spans. Span disagreements were resolved through majority-character aggregation and discussion-based adjudication.}{For span annotations, agreement was computed at the character level, yielding $\kappa$ = 0.55.}
% For span annotation, we adopted character-level majority aggregation: a character was labeled as hateful if selected by at least two annotators, and final spans were defined as maximal contiguous sequences of majority-selected characters. Multi-word spans were more frequent than single-token spans.

% \subsection{Disagreement Analysis and Resolution}

\subsection{\change{Disagreement Analysis and Resolution}{}}

\change{Beyond aggregate agreement, we analyse disagreement patterns across the three implicit-hate strategies. Per-strategy Fleiss' $\kappa$ values invert the frequency ordering: cultural/historical reference, the most common strategy (48\% of implicit-hate tweets), is also the most disagreement-prone ($\kappa = 0.43$), followed by sarcasm/figurative language ($\kappa = 0.55$) and other indirect strategies ($\kappa = 0.58$). Cultural-reference disagreements mainly reflected unequal recognition of political or historical allusions; sarcasm disagreements concerned whether figurative content was \emph{hate} or merely \emph{critical}. Implicit-hate rationale spans are also longer and more clause-level than explicit ones, and contain more named entities (43\% vs.\ 13\%), reflecting references to politicians and historical figures. Roughly 16\% of implicit-hate and 6\% of explicit-hate tweets required adjudication beyond majority voting, resolved through guideline-based discussion with unresolved ties defaulting to the majority label. Full per-strategy statistics appear in Appendix~\ref{sec:disagreement_full}.}{}

\subsection{Dataset Statistics and Analysis}
ParsHate consists of 10,000 Persian tweets collected between 2013 and 2022. Hate content constitutes 31\% of the dataset. As represented in Figure \ref{fig:r1}, temporal analysis reveals a steady increase in hate expression over the examined period. The proportion of tweets expressing hate rises from 5.90\% in the earliest year to a peak of 15.61\% in later years. 
With respect to explicitness, 7\% of hate tweets are annotated as implicit, suggesting that most hateful expressions in this corpus are overt. However, target expression exhibits a different pattern: 35\% of hate tweets contain implicit targets, where the attacked group is not directly named. Figure \ref{fig:r2} represents the distribution of implicitness for both hate and target over the years. Target distribution, which is heavily skewed toward politics, which accounts for 76.45\% of hate instances. Religious (6.42\%), nationality (4.90\%), occupation (3.45\%), ethnicity (2.29\%), gender (1.94\%), and other categories (4.55\%). Notably, 39\% of political hate instances involve implicit target references. Span-level analysis further shows that hate spans cover, on average, 18\% of tweet characters. Figure \ref{fig:r3} represents the distribution of targets over time.

\section{Analysis of ParsHate}
In this section, we analyze the temporal and structural properties of ParsHate. Specifically, we examine (i) temporal robustness of hate detection models across years; (ii) lexical variability across time using type–token ratio (TTR) and named entity masking; and (iii) multi-label target prediction behavior under class imbalance. Together, these analyses provide insight into the challenges posed by the dataset for hate and multi-label target prediction models.

\subsection{Hate-Speech Detection over Time}

To analyze temporal variation in ParsHate, we evaluate the generalization capacity of hate detection models across years (2013–2022). For each year, the data is randomly split into 80\% training and 20\% testing, and this split is kept fixed across all experiments to ensure comparability. 
\change{We fine-tuned both LLaMA-3 and Gemma-3; as Table \ref{tab:finetuned} shows, the two models yield nearly identical performance across all settings.}{}
%Following \cite{bokaei-etal-2025-culture}, and given LLaMA-3's marginally higher scores, we report the subsequent error analysis on LLaMA-3 as a representative case.}{}
All reported fine-tuned results use macro-F1 and are averaged over three random seeds, reported as mean ± standard deviation; zero-shot and SOTA baselines use single released checkpoints and therefore carry no seed variance.  We consider three training strategies to analyze temporal robustness:
(i) a PHATE-trained state-of-the-art (SOTA) LLaMA-3 model, trained using the original PHATE training split (2020–early 2023) \cite{Delbari_Moosavi_Pilehvar_2024} and evaluated on each ParsHate year. \change{PHATE and ParsHate share no overlapping tweets; for schema compatibility, although PHATE includes hateful, violent, and vulgar labels, we use only the \emph{hateful} label to align with ParsHate's binary hate/non-hate scheme;}{}
(ii) integrated training on 80\% of data from all ParsHate years (\textit{all-years}), followed by evaluation on each year separately; and
(iii) year-specific training, where models are trained and tested within the same year. For ParsBERT, year-specific training on the earliest years (2013–2016) failed to converge, collapsing to single-class predictions due to the limited and highly imbalanced per-year data; we therefore omit these cells (marked X in Table~\ref{tab:finetuned}) and report ParsBERT only where training was stable.

Tables~\ref{tab:zeroshot} and~\ref{tab:finetuned} present the performance of all experiments. In the zero-shot setting, the open models underperform (49.9 and 53.2 macro-F1 for LLaMA-3 and Gemma-3), while zero-shot GPT-5 is stronger (68.0); fine-tuning the open models yields large gains, confirming that the zero-shot gap is closed by supervision rather than reflecting a ceiling on the task. The integrated all-years training strategy achieves the highest overall performance (79.0 macro-F1 for LLaMA-3), consistently outperforming the PHATE-trained SOTA model (74.8) and year-specific models (63.5–67.7 across architectures). This suggests that temporally diverse training data improves robustness and mitigates year-specific distribution. The PHATE-trained SOTA model exhibits strong performance on temporally proximate years (2020–2022), reaching macro-F1 above 88, but performance declines substantially in earlier years (2013–2015), where scores fall to the mid-50s to mid-60s range. This degradation can reflect a temporal distribution shift, indicating that models trained on recent discourse struggle to generalize to earlier contexts. In contrast, integrated training across all years produces more stable performance across the temporal spectrum, reducing early-year degradation and improving average robustness. Year-specific training yields lower overall performance, suggesting that limited yearly data constrains generalization capacity. 

\begin{table*}[t]
\centering
\scriptsize
\setlength{\tabcolsep}{4pt}
\renewcommand{\arraystretch}{1.15}
\begin{tabular}{@{}c|ccc|ccc|ccc|ccc@{}}
\hline
& \multicolumn{9}{|c|}{Zero-shot}
& \multicolumn{3}{|c}{SOTA} \\
Year 
& \multicolumn{3}{|c|}{GPT-5}
& \multicolumn{3}{|c|}{LLaMA 3}
& \multicolumn{3}{|c|}{Gemma 3}
& \multicolumn{3}{|c}{\scriptsize{\cite{bokaei-etal-2025-culture}}} \\
 & P & R & F 
 & P & R & F 
 & P & R & F 
 & P & R & F \\
\hline
2013 & 54.6 & \textbf{66.2} & 59.9 & 35.4 & 32.2 & 33.8 & 32.1 & 35.5 & 33.7 & \textbf{65.4} & 51.7 & 57.9 \\
2014 & 51.4 & 62.7 & 56.4 & 37.1 & 33.8 & 35.4 & 42.6 & 32.3 & 36.7 & \textbf{68.2} & 49.5 & 57.3 \\
2015 & 50.8 & 50.4 & 50.6 & 44.5 & 41.2 & 42.8 & 52.2 & 48.8 & 50.4 & \textbf{69.6} & \textbf{61.3} & \textbf{65.2} \\
2016 & 66.7 & 62.3 & 64.4 & 38.4 & 45.9 & 41.8 & 47.6 & 44.4 & 45.9 & \textbf{72.5} & \textbf{74.6} & \textbf{73.5} \\
2017 & 58.6 & 56.9 & 57.7 & 51.8 & 55.6 & 53.6 & 58.5 & 51.4 & 54.7 & \textbf{73.8} & \textbf{65.2} & \textbf{69.2} \\
2018 & 68.9 & 72.3 & 70.6 & 52.3 & 57.7 & 54.9 & 62.3 & 55.1 & 58.5 & \textbf{75.4} & \textbf{77.8} & \textbf{76.6} \\
2019 & 76.8 & 74.7 & 75.7 & 56.4 & 54.1 & 55.2 & 51.6 & 60.8 & 55.8 & \textbf{82.6} & \textbf{88.3} & \textbf{85.3} \\
2020 & \textbf{88.5} & 76.9 & \textbf{82.3} & 55.7 & 55.2 & 55.5 & 60.2 & 64.4 & 62.2 & 78.7 & \textbf{84.1} & 81.3 \\
2021 & 67.9 & 71.5 & 69.7 & 58.3 & 62.7 & 60.5 & 63.9 & 68.2 & 66.0 & \textbf{93.4} & \textbf{88.6} & \textbf{90.9} \\
2022 & \textbf{93.6} & \textbf{87.8} & \textbf{90.6} & 68.6 & 62.2 & 65.2 & 67.3 & 69.8 & 68.5 & 91.3 & 85.7 & 88.4 \\
\hline
All & 67.8 & 68.2 & 68.0 & 49.9 & 50.1 & 49.9 & 53.8 & 53.1 & 53.2 & \textbf{77.1} & \textbf{72.7} & \textbf{74.8} \\
\hline
\end{tabular}
\caption{Precision (P), Recall (R), and F1 (F) scores across years and models for Zero-shot Experiment.}
\label{tab:zeroshot}
\end{table*}

% ===== TABLE 2: Fine-tuned (year-by-year + all-years) =====
\begin{table*}[t]
\centering
\scriptsize
\setlength{\tabcolsep}{3pt}
\renewcommand{\arraystretch}{1.15}
\resizebox{\textwidth}{!}{%
\begin{tabular}{@{}c|ccc|ccc|ccc|ccc|ccc@{}}
\hline
& \multicolumn{9}{|c|}{Fine-tuned \textit{year-by-year}} 
& \multicolumn{6}{|c}{Fine-tuned \textit{all-years}} \\
Year 
& \multicolumn{3}{|c|}{ParsBERT} 
& \multicolumn{3}{|c|}{LLaMA 3} 
& \multicolumn{3}{|c|}{Gemma 3} 
& \multicolumn{3}{|c|}{LLaMA 3} 
& \multicolumn{3}{|c}{Gemma 3} \\
 & P & R & F 
 & P & R & F 
 & P & R & F 
 & P & R & F
 & P & R & F \\
\hline
2013 & X & X & X & 42.5$_{\pm3.2}$ & 40.3$_{\pm3.4}$ & 41.4$_{\pm1.7}$ & 43.8$_{\pm3.3}$ & 45.6$_{\pm3.5}$ & 44.7$_{\pm1.8}$ & 64.3$_{\pm2.6}$ & 63.8$_{\pm2.8}$ & \textbf{64.0}$_{\pm1.4}$ & 63.2$_{\pm2.7}$ & 60.4$_{\pm2.9}$ & 61.7$_{\pm1.4}$ \\
2014 & X & X & X & 48.6$_{\pm3.5}$ & 42.4$_{\pm3.7}$ & 45.3$_{\pm1.9}$ & 55.7$_{\pm3.6}$ & 38.6$_{\pm3.8}$ & 45.5$_{\pm2.0}$ & 67.5$_{\pm2.8}$ & \textbf{66.3}$_{\pm3.0}$ & \textbf{66.9}$_{\pm1.5}$ & 67.3$_{\pm2.9}$ & 65.6$_{\pm3.1}$ & 66.4$_{\pm1.6}$ \\
2015 & X & X & X & 63.7$_{\pm3.9}$ & 59.8$_{\pm4.1}$ & 61.7$_{\pm2.1}$ & 60.4$_{\pm4.0}$ & 56.9$_{\pm4.2}$ & 58.6$_{\pm2.2}$ & 79.2$_{\pm3.1}$ & 75.5$_{\pm3.3}$ & 77.3$_{\pm1.7}$ & \textbf{80.4}$_{\pm3.2}$ & \textbf{78.1}$_{\pm3.4}$ & \textbf{79.2}$_{\pm1.8}$ \\
2016 & X & X & X & 60.5$_{\pm4.1}$ & 51.9$_{\pm4.3}$ & 55.9$_{\pm2.2}$ & 54.6$_{\pm4.2}$ & 56.2$_{\pm4.4}$ & 55.4$_{\pm2.3}$ & 71.6$_{\pm3.3}$ & \textbf{79.3}$_{\pm3.5}$ & \textbf{75.2}$_{\pm1.8}$ & 71.4$_{\pm3.4}$ & 78.3$_{\pm3.6}$ & 74.7$_{\pm1.9}$ \\
2017 & 61.3$_{\pm3.4}$ & 52.4$_{\pm3.6}$ & 57.2$_{\pm1.8}$ & 70.6$_{\pm3.3}$ & 62.8$_{\pm3.5}$ & 66.5$_{\pm1.8}$ & 60.9$_{\pm3.4}$ & 68.4$_{\pm3.6}$ & 64.4$_{\pm1.9}$ & \textbf{81.4}$_{\pm2.6}$ & 79.8$_{\pm2.8}$ & 80.6$_{\pm1.4}$ & 80.1$_{\pm2.7}$ & \textbf{82.3}$_{\pm2.9}$ & \textbf{81.1}$_{\pm1.5}$ \\
2018 & 64.3$_{\pm3.1}$ & 62.1$_{\pm3.3}$ & 63.2$_{\pm1.6}$ & 70.8$_{\pm2.9}$ & 68.4$_{\pm3.1}$ & 69.6$_{\pm1.6}$ & 74.2$_{\pm3.0}$ & 72.9$_{\pm3.2}$ & 73.5$_{\pm1.7}$ & 78.6$_{\pm2.2}$ & 80.4$_{\pm2.4}$ & \textbf{79.5}$_{\pm1.2}$ & 80.5$_{\pm2.3}$ & \textbf{82.1}$_{\pm2.5}$ & \textbf{81.3}$_{\pm1.3}$ \\
2019 & 59.1$_{\pm2.7}$ & 63.3$_{\pm2.9}$ & 61.1$_{\pm1.4}$ & 62.3$_{\pm2.6}$ & 64.5$_{\pm2.8}$ & 63.4$_{\pm1.4}$ & 66.9$_{\pm2.7}$ & 64.8$_{\pm2.9}$ & 65.8$_{\pm1.5}$ & \textbf{90.4}$_{\pm2.0}$ & 86.6$_{\pm2.2}$ & \textbf{88.4}$_{\pm1.1}$ & 89.3$_{\pm2.1}$ & 85.6$_{\pm2.3}$ & 87.4$_{\pm1.1}$ \\
2020 & 65.4$_{\pm2.9}$ & 69.5$_{\pm3.1}$ & 67.4$_{\pm1.5}$ & 67.8$_{\pm2.4}$ & 71.6$_{\pm2.6}$ & 69.6$_{\pm1.3}$ & 69.5$_{\pm2.5}$ & 73.4$_{\pm2.7}$ & 71.4$_{\pm1.3}$ & 77.5$_{\pm1.8}$ & 81.7$_{\pm2.0}$ & 79.5$_{\pm1.0}$ & 79.3$_{\pm1.9}$ & \textbf{83.3}$_{\pm2.1}$ & \textbf{81.4}$_{\pm1.0}$ \\
2021 & 78.4$_{\pm2.4}$ & 79.1$_{\pm2.6}$ & 78.7$_{\pm1.2}$ & 85.4$_{\pm2.2}$ & 77.6$_{\pm2.4}$ & 81.3$_{\pm1.2}$ & 82.6$_{\pm2.3}$ & 84.9$_{\pm2.5}$ & 83.7$_{\pm1.2}$ & \textbf{85.7}$_{\pm1.6}$ & \textbf{87.6}$_{\pm1.8}$ & \textbf{86.6}$_{\pm0.9}$ & 85.0$_{\pm1.7}$ & 81.2$_{\pm1.9}$ & 83.1$_{\pm0.9}$ \\
2022 & 79.0$_{\pm2.5}$ & 79.1$_{\pm2.7}$ & 79.1$_{\pm1.3}$ & 76.4$_{\pm2.0}$ & 82.5$_{\pm2.2}$ & 79.3$_{\pm1.1}$ & 79.3$_{\pm2.1}$ & 75.8$_{\pm2.3}$ & 77.5$_{\pm1.1}$ & 89.5$_{\pm1.4}$ & \textbf{91.2}$_{\pm1.6}$ & 90.3$_{\pm0.8}$ & \textbf{90.4}$_{\pm1.5}$ & 89.1$_{\pm1.7}$ & \textbf{89.7}$_{\pm0.8}$ \\
\hline
All & 68.0$_{\pm1.3}$ & 67.5$_{\pm1.4}$ & 67.7$_{\pm0.7}$ & 64.9$_{\pm1.3}$ & 62.2$_{\pm1.4}$ & 63.5$_{\pm0.7}$ & 64.8$_{\pm1.3}$ & 63.8$_{\pm1.4}$ & 64.3$_{\pm0.7}$ & \textbf{78.6}$_{\pm0.9}$ & \textbf{79.2}$_{\pm1.0}$ & \textbf{79.0}$_{\pm0.5}$ & \textbf{78.6}$_{\pm0.9}$ & 77.7$_{\pm1.0}$ & 78.2$_{\pm0.5}$ \\
\hline
\end{tabular}%
}
\caption{P, R, and F1  scores across years and models.  X marks unstable ParsBERT settings (see §4.1).}
\label{tab:finetuned}
\end{table*}

\begin{table*}[t]
\centering

\begin{minipage}[t]{0.36\textwidth}
\centering
\small
\setlength{\tabcolsep}{2.7pt}
\renewcommand{\arraystretch}{1.1}

\begin{tabular}{c|ccc|ccc}
\hline
& \multicolumn{3}{c|}{LLaMA-3} 
& \multicolumn{3}{c}{Gemma-3} \\
Hate-type & P & R & F1 & P & R & F1 \\
\hline
Implicit & 5.3 & 3.7 & 4.4 & \textbf{8.3} & \textbf{5.0} & \textbf{6.2}\\
Explicit & \textbf{84.6} & \textbf{84.4} & \textbf{84.5} & 84.3 & 83.1 & 83.7\\
\hline
Overall & \textbf{78.6} & \textbf{79.2} & \textbf{79.0} & \textbf{78.6} & 77.7 & 78.2\\
\hline
\end{tabular}

\caption{Models' performance according to the explicitness of hate. Models are trained on all years.}
\label{tab:explicit_implicit_yearly}
\end{minipage}
\hfill
\begin{minipage}[t]{0.60\textwidth}
\centering
\scriptsize
\setlength{\tabcolsep}{2pt}
\renewcommand{\arraystretch}{1.0}

\resizebox{\linewidth}{!}{
\begin{tabular}{l|ccc|ccc|ccc|ccc}
\hline
& \multicolumn{6}{c}{Unmasked} & \multicolumn{6}{c}{Masked} \\
\textbf{Target}
& \multicolumn{3}{c}{GPT-5}
& \multicolumn{3}{c}{LLaMA-3}
& \multicolumn{3}{c}{GPT-5}
& \multicolumn{3}{c}{LLaMA-3} \\
& P & R & F1 & P & R & F1 & P & R & F1 & P & R & F1 \\
\hline
Gender 
& \textbf{22.4} & \textbf{80.3} & \textbf{35.1} 
& 18.2 & 13.1 & 15.2 
& 21.3 & \textbf{80.6} & 33.7 
& 11.6 & 7.1 & 8.8 \\

Politics 
& 96.3 & 32.7 & 48.8 
& 84.6 & 96.2 & \textbf{90} 
& \textbf{97.4} & 27.6 & 43.0 
& 82.3 & \textbf{98.5} & \textbf{89.6} \\

Religion 
& 14.5 & \textbf{81.2} & 24.6 
& 65.1 & 32.7 & \textbf{43.6} 
& 14.7 & \textbf{81.4} & 24.6 
& \textbf{71.5} & 12.4 & 21.1 \\

Nationality 
& 7.3 & \textbf{60.7} & 13.4 
& 43.4 & 8.2 & 13.8 
& 8.4 & \textbf{60.5} & 14.8 
& \textbf{57.2} & 10.3 & \textbf{17.5} \\

Occupation 
& 5.5 & 60.4 & 10.1 
& 20.7 & 2.1 & 3.9 
& 8.2 & \textbf{65.7} & 14.6 
& \textbf{33.3} & 3.2 & \textbf{33} \\

Ethnicity 
& 4.7 & 60.1 & 8.8 
& \textbf{15.3} & 2.6 & 4.4 
& 6.6 & \textbf{65.2} & \textbf{12.1} 
& 14.2 & 2.3 & 4.0 \\

Other 
& 30.5 & 22.3 & 25.8 
& \textbf{46.3} & 6.3 & 11.1 
& 25.2 & \textbf{33.5} & \textbf{28.7} 
& 33.4 & 2.7 & 5.1 \\
\hline
Macro avg 
& 25.5 & 56.6 & 23.8 
& 42.0 & 23.0 & 26.0 
& 25.9 & \textbf{59.1} & 24.4 
& \textbf{42.8} & 19.5 & \textbf{27.1} \\
\hline
\end{tabular}
}

\caption{Per-target performance comparison across models.}
\label{tab:per_target_all}
\end{minipage}

\end{table*}

\subsection{Lexical Variability and NE Influence}
Given the observed performance degradation in the early years, we next examine whether lexical variability contributes to the shift in temporal distribution in ParsHate. To quantify lexical diversity, we compute the Type-Token Ratio (TTR) \cite{mcenery2011corpus}, defined as the ratio of unique words (types) to total words (tokens) in a given corpus. Before computation, tweets are normalized and tokenized using Hazm, a Persian tokenizer \footnote{https://pypi.org/project/hazm/0.9.1/}which has demonstrated strong performance for Persian text processing \cite{kamali2022evaluating}. 
We observed that the early years (2013–2015) exhibit the highest lexical diversity, followed by a gradual decline over time with minor fluctuations. Later years (2019–2022) show comparatively lower TTR values, indicating increased lexical stabilization in recent discourse. In the next step, to investigate whether named entities contribute to lexical variability, we perform an additional analysis by masking named entities using ParsBERT NER- as demonstrated in the Persian NER task, \cite{farahani2021parsbert}, and recomputing TTR. Although lexical diversity decreases after masking, early years remain consistently more diverse than later years. Notably, the strongest masking effect occurs in the earliest years (2013–2014), suggesting that named entities can contribute more to lexical variability during that period. From 2016 onward, the reduction in TTR after masking becomes smaller and stabilizes. 
Figure~\ref{fig:ttr} in the Appendix presents Lexical variability across years before and after masking.
% The overall pattern can be characterized as: early high diversity → gradual decline → mild fluctuation → late stabilization.
These findings indicate that early-year discourse is characterized by greater topical and lexical variation, partially driven by named entities. The higher lexical diversity in early years provides a plausible explanation for the reduced performance of temporally distant models observed in the previous subsection, as models trained on later, more stabilized discourse encounter broader lexical variation when applied to earlier data.

\subsection{\change{Multi-label Target Prediction}{Target Detection}}
\change{We formulate target prediction as a multi-label classification task: given a hate tweet, the model predicts which of the seven target categories (Gender, Politics, Religion, Nationality, Occupation, Ethnicity, Other) are attacked, allowing more than one label per tweet when multiple groups are targeted. This formulation follows the protocols of \citet{zampieri-etal-2019-predicting} and \citet{lu-etal-2023-facilitating}.}{We next examine multi-label target prediction performance.} As shown in Figure \ref{fig:r3}, political targets account for 76.45\% of hate instances, while other categories appear substantially less frequently.  This imbalance creates a realistic but challenging setting for multi-target modeling. We evaluate fine-tuned LLaMA-3 and GPT-5 for multi-target classification. As these models achieved the strongest performance in earlier hate detection experiments, we focus the multi-label target prediction analysis on them. We assess robustness by comparing performance before and after NE masking.

As represented in Table \ref{tab:per_target_all} LLaMA-3 demonstrates high precision for the Politics class (F1 = 0.89) but suffers from severe recall degradation on low-frequency targets such as Occupation and Ethnicity, resulting in near-zero F1 scores. In contrast, GPT-5 consistently achieves substantially higher recall for minority targets (often above 0.60), leading to improved F1 scores for rare categories along with low precision. To assess whether models rely on named entities for target prediction, we repeat the evaluation after masking named entities using ParsBERT \cite{farahani2021parsbert}. Masking reduces overall performance for both models as represented in Table \ref{tab:per_target_all} but preserves the fundamental behavioral contrast: LLaMA-3 remains dominant on Politics with high precision, while GPT-5 maintains broader recall across minority targets.  Overall, these findings highlight that ParsHate presents a structurally realistic yet challenging target distribution, where dominant political hate coexists with sparse identity-based categories. The contrasting behaviors of LLaMA-3 and GPT-5 further illustrate the tradeoff between precision and recall in low-resource target prediction settings. Using ParsHate's rationales as gold spans, we further evaluate whether models can \emph{localise} hate: LLaMA-3-FT reaches token-level span F1 of 44 overall (58 explicit, 14 implicit), with Gemma-3-FT showing the same pattern. This confirms that the rationale annotations serve not only as an interpretability resource but as a diagnostic of \emph{where} models fail; the full span-detection battery appears in Appendix~\ref{sec:span_detection_analysis}.

\subsection{Error Analysis}
To better understand model limitations under temporal diversity and structural imbalance, we conduct a detailed error analysis of the best-performing \textit{all-years} model (LLaMA-3-FT) and compare its behavior with GPT-5 as the next strong model. \change{LLaMA-3-FT and Gemma-3-FT achieve nearly identical overall performance under the \textit{all-years} setting (Table~\ref{tab:finetuned}), and, more importantly, exhibit the same error structure: both collapse on implicit hate while performing well on explicit hate (Table~\ref{tab:explicit_implicit_yearly}), and both show the same per-strategy difficulty ordering and span-localisation failure pattern (Appendices~\ref{sec:per_strategy_analysis} and~\ref{sec:span_detection_analysis}). Because the two models fail in the same way, not merely at the same rate, we report the detailed error analysis on LLaMA-3-FT as a representative case and use GPT-5 as a contrasting model.}{} We focus separately on hate detection and multi-label target prediction. Table \ref{fig:RQ1} in the Appendix represents some misclassification samples on these models.

\textbf{Hate Detection Error}: Although implicit hate constitutes only 7\% of hate instances in the dataset, it accounts for a large share of false negatives. In the \textit{all-years} model, implicit hate represents approximately 31\% of missed hate instances. Breaking down implicit errors by strategy reveals that cultural or historical references are the most challenging category, followed by sarcasm or figurative language, while the “other indirect” category contributes comparatively fewer errors. This suggests that instances requiring background knowledge and contextual interpretation are more likely to remain undetected than cases containing overt sentiment cues. In contrast, GPT-5 exhibits a lower concentration of implicit false negatives, with implicit hate accounting for approximately 22\% of its missed hate instances. However, GPT-5 introduces a higher number of false positives in figurative contexts. Overall, implicit rhetorical strategies—particularly those grounded in cultural or historical reference—constitute a large proportion of misclassification cases. Even under temporally integrated training.

\change{\textbf{Explicit vs.\ Implicit Performance Breakdown.} Using ParsHate's explicit/implicit annotations, we evaluate the two strongest fine-tuned models (LLaMA-3-FT and Gemma-3-FT) on explicit- and implicit-hate test subsets separately (Table~\ref{tab:explicit_implicit_yearly}). Both perform well on explicit hate, with average F1 = 84 for both models, but collapse on implicit hate, with F1 = 3 and F1 = 6, respectively. For LLaMA-3-FT, 83\% of implicit-hate errors are predicted as non-hate, 12\% as hate with the wrong target, usually \emph{politics}, and 5\% are partial-credit failures in multi-label cases. GPT-5 partly reduces this recall collapse and implicit hate accounts for 22\% of its missed hate cases versus 31\% for LLaMA-3-FT but over-flags neutral metaphor as hate in figurative contexts. Thus, neither model resolves implicit hate: LLaMA-3-FT mainly misses it, while GPT-5 improves recall at a precision cost. This suggests that background knowledge must be paired with the ability to distinguish hateful from neutral cultural references. Representative cases are provided in Appendix Table~\ref{tab:explicit_implicit_examples}.}{} 
\change{Full per-strategy implicit-hate analysis is provided in Appendix~\ref{sec:per_strategy_analysis}.}{}
\change{Span-detection performance are analysed in Appendix~\ref{sec:span_detection_analysis}.}{}

\textbf{Multi-label Target Prediction Error}: multi-label target prediction errors reveal that while implicit targets account for 35\% of hate instances in the dataset, they represent approximately 58\% of target misclassifications in the \textit{all-years} model. This confirms that target identification under indirect reference remains substantially more difficult than detecting the presence of hostility itself. LLaMA-3-FT exhibits that misclassified minority targets (e.g., occupation, ethnicity, gender) are frequently predicted as Politics, the majority class. This minority collapse reflects the model’s reliance on dominant target distributions when explicit identity cues are weak or absent. GPT-5  demonstrates a contrasting pattern. While it achieves higher recall for minority targets, it frequently over-assigns target categories, leading to lower precision. In several cases, GPT-5 predicts multiple plausible targets even when only one is supported by the text.

\change{\textbf{Qualitative Analysis of Annotator Rationales on Misclassifications.}
To complement the quantitative error breakdown, we examine annotator-selected rationale spans for misclassified implicit-hate and implicit-target cases in the \textit{all-years} model. Two representative cases illustrate distinct failure modes; the full set appears in Appendix Table~\ref{fig:rationale_full}.
In the first case, a tweet ending with ``what colour rope do you prefer, you British-bred mullahs?'' is annotated as implicit hate, with reference and sarcasm as rhetorical strategies. The hate is a culturally specific euphemism for execution, framed as a mock offer of choice. No token is hateful in isolation; recognising it requires knowledge that ``offering a rope'' encodes a call for hanging. The \textit{all-years} model predicts non-hate, illustrating the dominant implicit-hate failure mode: cultural or historical rationales convey hate only with external knowledge.The second case shows a failure mode tied to implicit targets. In a tweet mourning \textit{Mahsa Amini} a widely publicised case and focal point of political protest in Iran in 2022, annotators highlighted spans including ``bastards'' and ``what did you do to this girl,'' reflecting hostility toward the regime. The model detects hate but assigns the target as \emph{gender}, based on the surface mention of ``this girl.'' Here the hate is explicit, but the target is implicit and discourse-level: the addressees are state actors, not a gender group. This shows why implicit-target identification requires reasoning beyond local lexical signals. Across the full set, rationales show where cultural, historical, or rhetorical knowledge is needed and where the model lacks signal. They serve both as supervision signals and as diagnostic evidence of the background knowledge current models fail to capture.}{} Implicitness is a central source of modelling difficulty in ParsHate: even strong models struggle when hate is indirect or targets are underrepresented. Together with temporal variation, lexical diversity, and class imbalance, it shapes the dataset’s core challenges.

% Together, these findings show that implicitness is a central driver of modelling difficulty in ParsHate. Even strong models struggle to resolve identity categories when hostility is indirect or targets are underrepresented.

% Overall, we observed that temporal variation, lexical diversity, class imbalance, and implicitness jointly shape the core modelling challenges in ParsHate.

% These analyses show that ParsHate presents structural and temporal challenges for Persian hate and multi-label target prediction. Temporal evaluation reveals distribution shift, with models trained on recent discourse struggling to generalise to earlier years with higher lexical variability. Lexical analysis shows that early years are more diverse, only partly due to named entities, suggesting broader topical variation over time. Structurally, severe target imbalance, especially the dominance of political hate, biases models toward majority-class predictions. Error analysis further shows that implicit hate and implicit targets account for many misclassifications. Overall, temporal variation, lexical diversity, class imbalance, and implicitness jointly shape the core modelling challenges in ParsHate.

\section{Conclusion}

We introduce ParsHate, the first decade-spanning (2013--2022) Persian benchmark for hate and multi-label target prediction, with 10,000 manually annotated tweets, a structured target taxonomy, explicit/implicit distinctions, strategy categories, and span-level rationales. Score-stratified temporal sampling preserves natural label distributions while reducing keyword-driven bias. Evaluation reveals both temporal distribution shift across years and structural collapse on implicit hate and minority targets.  Addressing these challenges requires better handling of culturally grounded hate and rare target categories, highlighting important directions for Persian hate-speech detection.

\section*{Ethics Statement}
ParsHate is built from a public archival sample of Twitter; we release only task-relevant tweet text with user identifiers removed and do not attempt to identify any individual. Annotation was carried out by three native Persian speakers who were informed the data contains distressing content, could opt out at any time, and were compensated at a fixed rate. ParsHate is released to support research on hate-speech and target detection in Persian, a low-resource language. Because it contains offensive and potentially distressing content and annotates the targets of hate, it could be misused; we therefore release it for research purposes only, ask that it be treated as sensitive, and release no information that could re-identify individuals.

\section*{Limitation}
First, although the dataset spans ten years, it is restricted to Twitter and may not generalize to other Persian-speaking platforms such as Instagram or Telegram, where discourse norms and interaction patterns differ. Second, the dataset exhibits substantial class imbalance, particularly the dominance of political hate. While this reflects naturally occurring discourse, it constrains model performance on minority target categories and limits balanced evaluation across identity dimensions. Moreover, target classification performance remains considerably lower than binary hate detection, especially for low-frequency identity groups. This highlights the structural difficulty of multi-target modeling under severe imbalance and implicit reference. A more comprehensive investigation of target-specific modeling strategies, balancing techniques, and architectural adaptations lies beyond the scope of the present work and remains an important direction for future research. Third, implicit hate constitutes a relatively small proportion of the dataset (7\%), which restricts fine-grained modeling and detailed statistical analysis of implicit rhetorical strategies. Although we provide error analysis for both hate and multi-label target prediction, a deeper qualitative investigation would further illuminate model failure patterns. Finally, all experiments focus on supervised evaluation within ParsHate and cross-temporal transfer from PHATE. Broader cross-domain, cross-platform, and cross-cultural generalization remain to be explored. Future work may expand implicit hate coverage, incorporate additional platforms, and conduct more extensive qualitative analyses of implicit rhetorical strategies.

% \section*{Acknowledgements}

\bibliography{anthology,custom}
\bibliographystyle{acl_natbib}
% \bibliography{ref.bib}

\appendix

\section{Appendix}
\label{sec:appendix}

\subsection{\change{Per-Strategy Analysis of Implicit Hate}{}}
\label{sec:per_strategy_analysis}
This section expands on the per-strategy implicit-hate analysis referenced in Section~4.4. LLaMA-3-FT's F1 on implicit hate decomposes by rhetorical strategy as follows: cultural/historical reference F1 = 2, sarcasm/figurative language F1 = 4, other indirect strategies F1 = 5. Gemma-3-FT exhibits the same ordering across strategies, indicating that the gradient is a robust cross-model pattern. All three values are far below the explicit-hate F1 of 84.

This gradient mirrors the type and weight of external knowledge each strategy demands. Cultural and historical references encode hostility entirely in the referent, specific events, public figures, or political symbols, with no intrinsic textual cue; recognition requires Persian-specific event and entity knowledge. Sarcasm and figurative language retain partial structural cues such as negation, exaggerated affect, or dehumanising comparison, which enable occasional detection but are systematically defeated when the irony depends on contradiction with extra-textual context. The other-indirect category is comparatively the most tractable, as it typically retains some lexical hostility marker, exclusionary phrasing, generalisation-via-quantifier, or collective accusation, that is localisable from the text alone. The bottleneck is therefore not only implicitness per se but the knowledge load that each strategy imposes.

Notably, cultural/historical reference is both the hardest strategy and the most frequent in ParsHate, accounting for 48\% of implicit-hate instances (versus 36\% for sarcasm/figurative and 16\% for other indirect). The model fails most precisely on the strategy that dominates the implicit-hate distribution. This skew toward historical and political reference is itself a Persian-specific finding: it reflects the structure of Persian online discourse, which is heavily anchored in shared political and religious memory, and is broadly distinct from the strategy distributions reported in English implicit-hate research \citep{elsherief-etal-2021-latent}.

Three representative cases, ordered from text-localisable to maximally knowledge-dependent, illustrate this gradient (LLaMA-3-FT and Gemma-3-FT both predict non-hate for all three):

\textbf{(a) Other indirect.} ``Those people aren't Iranian; they all act for the internet-filtering government.'' Hostility is conveyed through identity denial, excluding a group from the national identity, and collective accusation, without slur, metaphor, or named referent. The exclusionary phrasing is localisable from the surface text alone.

\textbf{(b) Sarcasm/figurative.} ``When people like Zarif wag their tails like loyal dogs for the U.S., what is left to say?'' The expression uses animalisation as a sarcastic label; in Iranian cultural context, calling someone a dog is a strong insult implying low status and servitude, constituting dehumanisation. The structural cue (the dehumanising comparison) is text-internal, but the hostile reading depends on culturally specific affective weight.

\textbf{(c) Cultural/historical reference.} ``Now they'd better not say anything about religion or faith anymore, when they hug Ch\'avez's mother, this whole crowd is all the same!'' The hostility depends on knowing the politically contested moment when Mahmoud Ahmadinejad embraced Hugo Ch\'avez's mother at the latter's funeral, an event that became contested in Iranian religious discourse. Without this background, the tweet reads as a neutral generalisation.

\subsection{\change{Span-Detection Analysis}{}}
\label{sec:span_detection_analysis}
Beyond tweet-level classification, ParsHate's span-level rationales enable fine-grained evaluation of whether models can \emph{localise} hate. Following the toxic-spans paradigm of \citet{pavlopoulos-etal-2021-semeval}, with related span-level hate-speech evaluations in HateXplain \citep{mathew2021hatexplain} and STATE ToxiCN \citep{bai-etal-2025-state}, we evaluate span identification on all gold-hate tweets using token-level F1, with character-level annotations converted via Hazm tokenisation consistent with Section~4.2.

On LLaMA-3-FT, overall span F1 = 44, decomposing into F1 = 58 on explicit hate and F1 = 14 on implicit hate. Gemma-3-FT shows the same pattern. The 58 $\rightarrow$ 14 gap reflects a qualitative shift in failure type, not merely a quantitative drop. On explicit hate, errors are predominantly \emph{boundary errors}: the model identifies the hate region but selects a span that is too narrow, too broad, or fragmented relative to the gold annotation. Partial overlap still earns partial F1 credit, which is why explicit F1 lands at 58 rather than near zero. On implicit hate, errors are predominantly \emph{localisation errors}: the model either fails to produce a span at all or marks a non-anchor portion of the text. These produce zero or near-zero overlap, dragging F1 down to 14.

Decomposing the failure modes for misclassified implicit-hate span predictions on LLaMA-3-FT: 80\% are no-span (the model declines to localise, mirroring the 83\% non-hate prediction rate on implicit hate reported earlier in Section~4.4), 12\% mark the wrong anchor (typically a named entity or emotionally charged surface token rather than the rhetorical device that carries the hostility), 5\% mark the whole tweet (an over-generation default), and 3\% are other partial-overlap cases. The dominance of the no-span mode confirms that span-detection failure is downstream of detection failure: if the model does not recognise hate, it has nothing to localise.

Two cases illustrate the contrast between failure modes (gold spans and model predictions translated for presentation):

\textbf{(i) Explicit hate, boundary error.} Tweet: ``Each time they leave a family in mourning, drag a family behind ICU doors, behind prison walls, into courtroom corridors, into darkness and grief, killing people, women and men alike\ldots{} Damn it, you sons of dogs, where did you come from to be this animalistic?'' The gold annotation covers two non-contiguous spans: (a) the full enumeration of hate actions, and (b) the closing slur-plus-dehumanisation clause. LLaMA-3-FT marks only three short fragments inside (a) (``leave a family in mourning'', ``behind prison walls'', ``killing people'') and misses (b) entirely. This is a classic under-coverage failure where the model locates roughly where hate is present but cannot capture the full rhetorical arc, and under-counts when multiple non-contiguous hate spans co-occur.

\textbf{(ii) Implicit hate, localisation error.} Tweet: ``But honestly, when Raisi gets emotional and speaks angrily, he really gives off a sense of killing.'' Hate is encoded entirely in the rhetorical phrase \emph{``gives off a sense of killing''} (a colloquial Persian construction implying menace and dehumanisation). The gold span is exactly this clause. LLaMA-3-FT produces no span at all, it does not register the rhetorical anchor as hate because, taken literally, the phrase contains no overt hostility marker.

The span-detection failure on implicit hate provides direct evidence at sub-tweet granularity: even when the input is restricted to known hate tweets, i.e., the model is implicitly told ``hostility is present, find it'', localisation still fails approximately 80\% of the time. The hate is not in any localisable token; it is in the relationship between the text and an external referent (an event, a public figure, or a culturally entailed implication). Therefore the annotated rationales in ParsHate are therefore not only an evaluation resource but a potential supervision signal for future rationale-aware or knowledge-augmented training.

\subsection{\change{Annotator Demographics}{}}
All annotations were carried out by three female native Persian speakers, aged 25--35, who each lived in Iran for over twenty years. All hold graduate degrees (two in mathematics, one in computer science). Annotators were compensated at a fixed rate. They were informed before the task that the dataset contains hateful and potentially distressing content, were free to take breaks at any point, and could opt out at any stage. The annotation guidelines (approximately five pages, prepared in Persian; English translation in the supplementary materials) and annotation interface will be released publicly with the dataset.

\subsection{\change{Full Disagreement Statistics}{}}
\label{sec:disagreement_full}

This section expands on the disagreement analysis summarised in Section~3.3.

\textbf{Per-strategy frequencies and disagreement rates.} Among the three implicit strategies, cultural or historical reference is the most common in ParsHate (48\% of implicit-hate tweets) and the most disagreement-prone (38\% of cases), followed by sarcasm or figurative language (36\% frequency, 27\% disagreement) and other indirect strategies (16\% frequency, 22\% disagreement). Per-strategy Fleiss' $\kappa$ values show the inverse trend: $\kappa = 0.43$ for cultural/historical reference, $0.55$ for sarcasm, and $0.58$ for other indirect strategies.

\textbf{Rationale-span structure.} Implicit-hate spans are longer than explicit ones (31\% vs.\ 17\% of tweet characters) and predominantly clause-level: 57\% multi-token clauses, 29\% noun phrases, and 14\% mixed or multi-clause selections. Explicit spans, by contrast, rely more on short hate lexicon. Named entities appear in 43\% of implicit spans versus 13\% of explicit ones, reflecting references to politicians, regimes, and historical figures.

\textbf{Adjudication procedure.} About 16\% of implicit-hate tweets and 6\% of explicit-hate tweets required adjudication beyond majority voting. Resolution was two-step: multi-label assignment was allowed when multiple rhetorical strategies plausibly applied; remaining disagreements, typically on the implicit/explicit boundary or whether figurative language qualified as hate, were discussed using the guidelines and pilot examples, with unresolved ties defaulting to the majority label.

% \clearpage

\begin{table*}[t]
\centering
\includegraphics[width=0.9\textwidth]{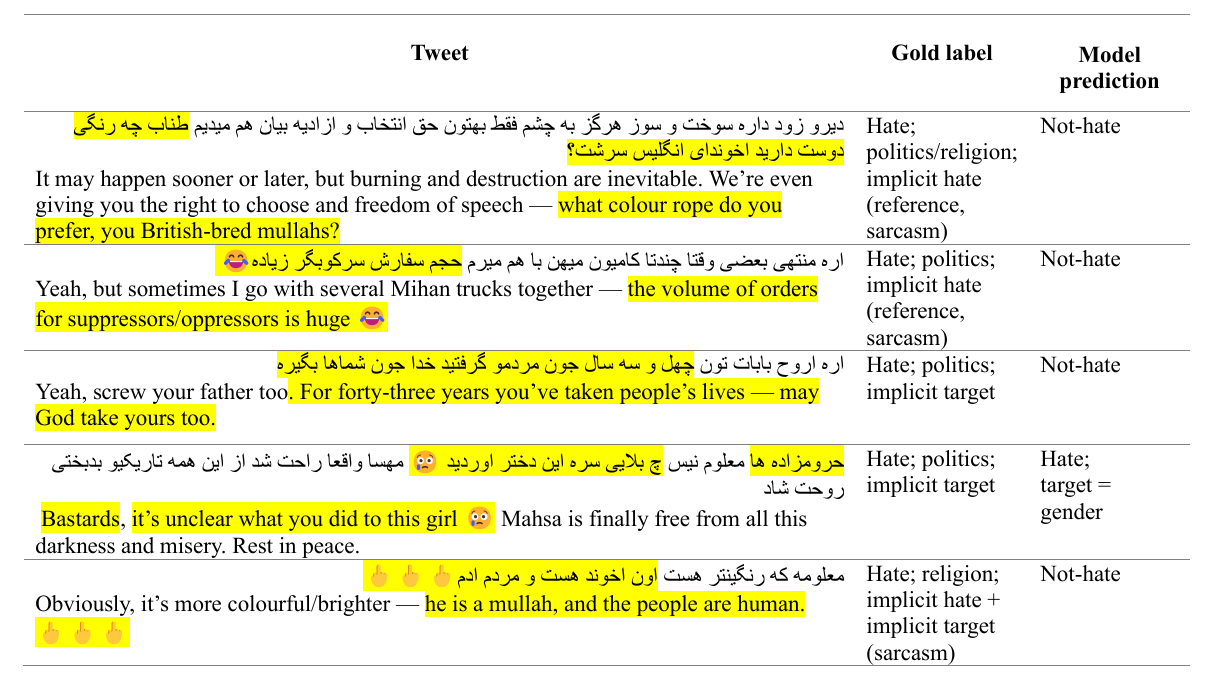}
\caption{\change{Annotator rationale spans on representative \textit{all-years} misclassifications.}{}}
\label{fig:rationale_full}
\end{table*}

\begin{table*}[t]
\centering
\includegraphics[width=0.9\textwidth]{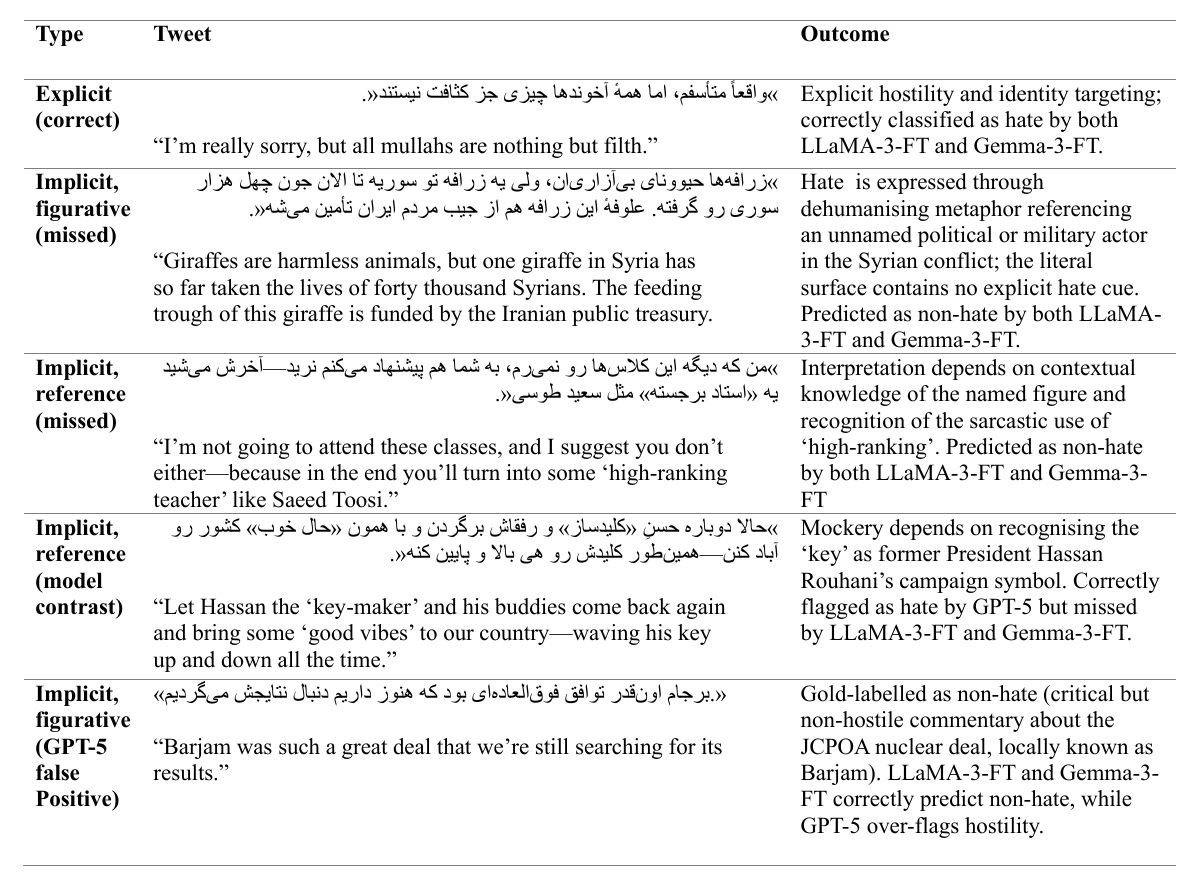}
\caption{\change{Representative cases illustrating the explicit--implicit performance gap}{}}

\label{tab:explicit_implicit_examples}
\end{table*}

\begin{figure*}[t]
\centering
\includegraphics[width=0.8\textwidth]{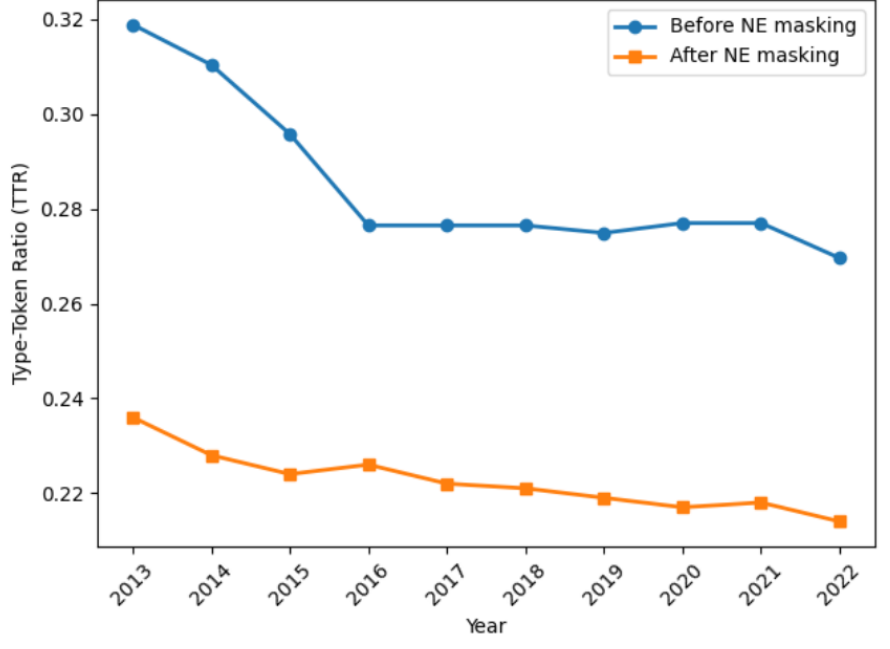}

\caption{Lexical variability (TTR) before and after named-entity masking across years.}
\label{fig:ttr}
\end{figure*}

\begin{table*}[t]
\centering
\includegraphics[width=0.9\textwidth]{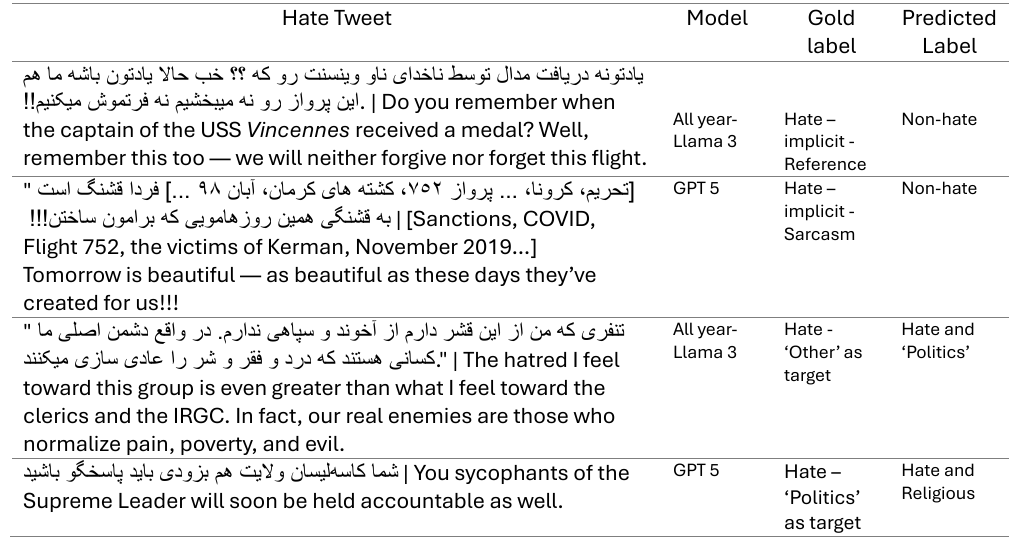}
\caption{Misclassification samples for the best models.}
\label{fig:RQ1}
\end{table*}

% \begin{figure*}[t]
% \centering

% \begin{minipage}[t]{0.48\textwidth}
% \centering
% \includegraphics[width=\linewidth]{ttr2.pdf}
% \caption{Lexical variability (TTR) before and after named-entity masking across years.}
% \label{fig:ttr}
% \end{minipage}
% \hfill
% \begin{minipage}[t]{0.48\textwidth}
% \centering
% \includegraphics[width=\linewidth]{misclassification1.pdf}
% \caption{Misclassification samples for the best models.}
% \label{fig:RQ1}
% \end{minipage}

% \end{figure*}

\clearpage

\subsection{\change{Annotation Guidelines}{}}
\label{sec:annotation_guidelines}

The full annotation guidelines provided to annotators are reproduced below in English translation; the original Persian version will be released publicly with the dataset.The aim of these guidelines is to provide a structured procedure for identifying and categorising hate speech in Persian tweets. They define clear criteria for distinguishing hate speech from violent or merely offensive content, explain how to identify the targeted individual or group, and describe how each target is assigned to a predefined category. They also distinguish between explicit and implicit hate speech and provide key cues for recognising indirect forms of hate, such as figurative language or references to historical and cultural events or symbols. Annotators are expected to apply these guidelines carefully and consistently.

\paragraph{Step 1 -- Identifying the Presence of Hate Speech.}

A tweet should be labelled as containing hate speech if it expresses or encourages hatred toward an individual or group on the basis of their affiliation with a specific category such as political views, occupation, religion, nationality, ethnicity, gender, age, disability, or similar.

\textit{Distinguishing hate speech from violence and offensive language.} It is important to distinguish hate speech from violent or offensive content. While such content can be harmful, it does not always fall within the definition of hate speech. In other words, a tweet may contain violent or offensive content without being hateful.

\textbf{Violent content} includes the following:
\begin{itemize}\itemsep0pt
\item Threats of violent acts against a specific individual or group.
\item Expressing wishes, hopes, or desires for the death or serious physical harm of others, or inciting such outcomes.
\item Inviting or encouraging others to harass an individual or group.
\end{itemize}

\textit{Example.} ``We will make your living and dead bodies tremble; this time you won't escape death, you dug your own grave.'' This tweet contains violence (a threat of violent action against a specific group) but does not contain hate speech.

\textbf{Offensive language} includes the use of vulgar or profane language without targeting a group based on identity, or insulting a specific individual without framing them as a representative of a broader group.

\textit{Example.} ``This Mahdi Tootonchi guy is putting on an act, or he really is just that idiotic.'' The tweet uses only vulgar and offensive language; there is no hate speech.

Note that hate speech can overlap with violent or offensive language, but not every violent or offensive expression constitutes hate speech.

\textit{Example.} ``Bastards, you executed people to scare us? Nothing matters to me anymore, I'll stain my hands with your filthy and impure blood.'' This contains \emph{hate speech} (for describing the hateful act of execution), is also labelled as \emph{offensive} (for the vulgar term ``bastards''), and contains \emph{violence} (for ``I'll stain my hands with your blood'').

\paragraph{Step 2 -- Explicit vs.\ Implicit Hate Speech.}
If a tweet conveys hate indirectly, using non-literal language, it is classified as \emph{implicit hate speech}. Such language typically avoids overt hate vocabulary. When a tweet is labelled as implicit hate, the \emph{strategy} by which the hate is conveyed must also be specified (figurative language, historical/cultural reference, or other indirect means). Three strategies are considered for implicit hate:

\textbf{(i) Figurative language:} the use of symbolic and non-literal expressions to convey hate, such as irony, sarcasm, or exaggerated comparisons.

\textit{Example.} ``These people are stuck to the country's resources like leeches again.'' A group of people is figuratively compared to leeches, bloodsuckers, implicitly conveying exploitation, parasitism, or harm to society. The hate is carried without any explicit hate vocabulary.

\textit{Example.} ``Hassan the `key-maker' has been so clever and precise in making his keys that the elites are fleeing the country one by one! Well done!'' On the surface, the sentence describes Hassan as a clever and meticulous key-maker, but the tone is clearly mocking. The speaker sarcastically attributes the emigration of elites to his performance, indirectly invoking his incompetence. Sarcasm allows the speaker to convey hostility or anger without using overt hate vocabulary; this makes the sentence a form of implicit hate speech in which the negative message is delivered through irony.

\textbf{(ii) Reference to historical or cultural events or symbols:} the use of events, figures, or symbols associated with hatred to convey a hateful message without expressing it directly.

\textit{Example.} ``We don't want them to repeat November 2019 for our young people again, right?'' Without naming any individual or group, the speaker refers to a historical event (November 2019) that, in public memory, is associated with repression, protests, or state violence. The reference implicitly conveys warning, threat, or distrust, all of which contribute to hate, toward a political group or governing body. Such historical or cultural references can indirectly evoke hate or political/social opposition and produce a polarising atmosphere, even when the surface of the sentence appears supportive or restrained.

\textbf{(iii) Other:} any other indirect means of conveying hate that does not fall under figurative language or historical/cultural reference.

\textit{Note.} A tweet labelled as implicit hate may use more than one strategy simultaneously, in which case multiple options may be selected. For example, the ``Hassan the key-maker'' tweet above uses both figurative language and historical/cultural reference (alluding to former President Hassan Rouhani's display of a key during one of his interviews).

\paragraph{Step 3 -- Identifying the Target of Hate Speech.}
If a tweet is identified as hate speech, the next step is to determine the targeted individual or group. The target can fall into one of the following predefined categories:

\textbf{Gender.} Hate directed at individuals or groups on the basis of their gender or gender identity.

\textit{Example.} ``It was these same women who, by going around without their proper clothing, turned everyone else into trash! Get lost, and now they also want the right to divorce and whatnot!''

\textit{Note.} If gender is the target, specify whether the target is \emph{women} or \emph{men}.

\textbf{Politics.} Hate directed at political ideologies, parties, politicians, or political supporters.

\textit{Example.} ``Folks, do you hear our children's voices from Evin? Do you realise that if this regime lasts another year, we'll essentially no longer hear the name Iran?''

\textit{Note.} If politics is identified as the target, specify whether the target is \emph{domestic} or \emph{foreign}. The tweet above targets the Iranian regime (domestic). In contrast, ``Monarchists residing in England have no right to demand! For years they have been making our situation worse and fishing in muddy waters!'' targets monarchists residing in England (foreign).

\textbf{Religion.} Hate directed at individuals or groups because of their religious beliefs or affiliations.

\textit{Example.} ``So statues are forbidden in the clergy's interpretation of Islam, but when it comes to their own criminals it suddenly becomes permissible, they truly do worship false idols!''

\textit{Note.} If religion is the target, specify whether the target is \emph{Islam} or \emph{another religion}. In the example above the target is Islam.

\textbf{Nationality.} Hate expressed on the basis of country of origin or citizenship.

\textit{Example.} ``Do you get it that Afghans have ruined Iran?''

\textit{Note.} If nationality is the target, specify which nationality is targeted (Iranian, Afghan, Arab nations, Western nations, or other).

\textbf{Occupation.} Hate that targets people because of their profession or occupational role.

\textit{Example.} ``I've never seen a more rotten group than doctors in Iran! Freeloaders and rude! Total scum!''

\textbf{Ethnicity/Race.} Hate based on race, ethnicity, or perceived family background.

\textit{Example.} ``Even with private tutoring, when a Lor keeps failing, honestly, education isn't their fault, they really don't get it!''

\textbf{Other.} If the target of hate does not fit any of the categories above, the \emph{Other} option is selected.

\textit{Note.} A tweet may contain multiple targets; in that case all applicable options are selected.

\paragraph{Step 4 -- Explicit vs.\ Implicit Targets.}
As in Step~2, annotators indicate whether the target of hate is mentioned directly or invoked indirectly through non-literal language.

\textit{Example.} ``The tortured and pain-stricken bodies, the spilled blood at the hands of \#Zahhak, are the price paid for freedom, no room for grief!!! Until we are struck down and martyrs are made, there is no freedom. We have no time to mourn, this is the time for war!'' Instead of directly naming a specific person, the word \emph{Zahhak} (a mythological tyrant in Persian literature) is used.

\textit{Example.} ``These bastards I hate, they made it to parliament again!'' A pronoun (\emph{these bastards}) is used instead of directly naming the hated individuals.

\paragraph{Step 5 -- Identifying the Hate Span.}
Annotators mark all the most minimum yet specific parts of the tweet where the hate is expressed using square brackets~[\,].

\textit{Example.} ``In every language, declare [the IRGC a terrorist organisation] and demand [the expulsion of the regime's ambassadors].''

\textit{Example.} ``These same religious people who, the moment something is forbidden and illegal, oppose it and [tear apart whoever does that thing], the moment that thing becomes permissible for any reason, they themselves are first in line to do it. They [don't oppose anything on rational grounds]; their only justification is its being forbidden [a clear example is that they were the first to send their wives to the corrupt and morally compromised sports stadium environment].''

\paragraph{Recording the Annotations.}
All responses are recorded as binary values (0 or 1), where 1 indicates presence or an affirmative response and 0 indicates absence or a negative response.

\paragraph{Summary of the Labelling Procedure.}

\begin{enumerate}\itemsep0pt
\item Does this tweet contain hate speech? \emph{If not, stop.}
\item Is the hate expressed implicitly or explicitly (directly or indirectly)?
\item If implicit, specify the strategy (figurative language, historical/cultural reference, or other).
\item Specify the target (gender, politics, religion, occupation, nationality, ethnicity, or other), with the relevant subcategory (e.g., women/men; domestic/foreign; Islam/other; Iranian/Afghan/Arab/Western/other).
\item Is the target expressed implicitly?
\item Mark the hate span(s).
\end{enumerate}

\end{document}